\documentclass[sigconf]{acmart}

\usepackage{booktabs}
\usepackage{graphicx}
\usepackage{multirow}
\usepackage{array}
\usepackage{url}
\usepackage{xspace}
\usepackage{amsmath}
\usepackage{balance}
\newcommand{\MISR}{M$^3$ISR\xspace}

\AtBeginDocument{%
  }

\setcopyright{acmlicensed}
\copyrightyear{2026}
\acmYear{2026}
\setcopyright{cc}
\setcctype{by}
\acmConference[RichMediaGAI '26]{The fourth International Workshop on Rich Media with Generative AI }{November 10--14, 2026}{Rio de Janeiro, Brazil}
\acmBooktitle{The fourth International Workshop on Rich Media with Generative AI (RichMediaGAI '26), November 10--14, 2026, Rio de Janeiro, Brazil}
\acmDOI{10.1145/3841458.3841537}
\acmISBN{979-8-4007-2947-8/2026/11}

\begin{document}

\title{M$^3$ISR: A Multi-Modal Multi-View Benchmark \\
for 3D/4D Gaussian Splatting and Feedforward Compression}

\author{Xinhui Liu}
\affiliation{%
  \institution{The University of Hong Kong}
  \city{Hong Kong}
  \country{China}}
\email{xhliu01@hku.hk}

\author{Lei Liu}
\affiliation{%
  \institution{The University of Hong Kong}
  \city{Hong Kong}
  \country{China}}
\email{liulei95@hku.hk}

\author{Zhenghao Chen}
\affiliation{%
  \institution{University of Newcastle}
  \city{Callaghan}
  \state{NSW}
  \country{Australia}}
\email{zhenghao.chen@newcastle.edu.au}

\author{Lebin Zhou}
\affiliation{%
 \institution{Santa Clara University}
 \city{Santa Clara}
  \state{CA}
  \country{USA}}
\email{lzhou@scu.edu}

\author{Wei Wang}
\affiliation{%
  \institution{Futurewei Technologies, Inc.}
  \city{San Jose}
  \state{CA}
  \country{USA}}
\email{rickweiwang@futurewei.com}

\author{Wei Jiang}
\affiliation{%
  \institution{Futurewei Technologies, Inc.}
  \city{San Jose}
  \state{CA}
  \country{USA}}
\email{wjiang@futurewei.com}

\renewcommand{\shortauthors}{Xinhui Liu et al.}

\begin{abstract}
High-fidelity free-viewpoint video (FVV) and interactive rendering increasingly rely on explicit Gaussian representations, yet practical deployment remains constrained by representation size, dynamic updates, and computational cost. Existing multi-view video benchmarks provide valuable real-captured content, but they make it difficult to isolate the effects of controlled camera geometry, representation efficiency, and temporal redundancy. We introduce M$^3$ISR, a controlled synthetic benchmark for 3D and 4D Gaussian Splatting (3DGS/4DGS). The benchmark contains 25 scenes from five indoor and outdoor scene groups, two camera/motion configurations, six synchronized 1080p views, and dense ground-truth annotations including RGB, camera parameters, depth, semantic and instance segmentation, and static--dynamic masks. The shared-center camera design intentionally isolates angular view variation and enables controlled evaluation of novel-view synthesis and representation efficiency. We organize M$^3$ISR into five complementary tracks covering 3DGS synthesis, 4DGS synthesis, 4DGS streaming, 3DGS compression, and 4DGS compression. Representative baseline results show small differences in static reconstruction quality but substantial differences in representation storage, while the evaluated streaming methods exhibit substantially higher reported training or reconstruction cost than the corresponding offline dynamic reconstruction baselines. We further define feedforward compression tasks for 3DGS and 4DGS and provide reference rate--distortion formulations and preliminary baseline evaluations. The benchmark is intended as a controlled and complementary testbed for systematic study of Gaussian-based FVV reconstruction, compression, and streaming.
\end{abstract}




\maketitle

\section{Introduction}
Free-viewpoint video (FVV) aims to synthesize photorealistic visual content from viewpoints beyond the captured cameras, and has become an important building block for immersive media, telepresence, extended reality, and interactive 3D applications. Recent advances in neural scene representations, particularly 3D Gaussian Splatting (3DGS)~\cite{kerbl20233d}, have made high-quality novel-view synthesis practical with efficient rasterization. Its dynamic extensions, including deformable Gaussian representations and 4D Gaussian Splatting (4DGS)~\cite{yang2024deformable,wu20244d}, further enable free-viewpoint video generation from time-varying scenes.

Despite this progress, existing benchmarks only partially support the systematic evaluation required for practical FVV systems. Static datasets such as Mip-NeRF 360~\cite{barron2022mip} primarily evaluate novel-view reconstruction, while dynamic resources such as D-NeRF and multi-view video datasets introduce temporal content but do not explicitly target the joint evaluation of FVV generation and delivery efficiency. On the other hand, image-domain video compression benchmarks such as Vimeo-90K focus on bitrate--fidelity trade-offs without providing explicit 3D scene geometry, synchronized multi-view observations, or camera calibration.

To bridge these gaps, we introduce \MISR, a fully synthetic and richly annotated benchmark designed for systematic evaluation of both \emph{generative FVV synthesis} and \emph{efficient FVV delivery}. \MISR extends the multi-view benchmark paradigm toward Gaussian-based representations by providing static environments, dynamic objects, synchronized ego-centric cameras, and dense ground-truth annotations. Each scene is rendered from six synchronized cameras under a controlled shared-center geometry, together with RGB images, calibrated camera parameters, depth, semantic and instance segmentation, and static--dynamic scene annotations. This controlled design enables reproducible evaluation of spatial fidelity, temporal consistency, representation compactness, and end-to-end delivery efficiency.

The benchmark is designed around two complementary objectives. The first is \emph{generative FVV synthesis}, which evaluates whether 3DGS- and 4DGS-based methods can recover high-fidelity novel views and maintain temporal consistency under sparse multi-view observations. The second is \emph{efficient FVV delivery}, which evaluates whether these representations can be compressed, transmitted, decoded, and rendered under practical storage, bitrate, and latency constraints.

Accordingly, we organizes \MISR into five tracks: Track 1 for 3DGS-based FVV synthesis, Track 2 for 4DGS-based FVV synthesis, Track 3 for 4DGS streaming, Track 4 for 3DGS compression, and Track 5 for 4DGS compression. Together, the five tracks cover the full pipeline from scene reconstruction and dynamic view synthesis to representation compression and low-latency delivery.

Our contributions are as follows:
\begin{itemize}
  \item \textbf{A controlled multi-view FVV benchmark.}
  \MISR provides 25 synthetic indoor and outdoor scenes, two camera/motion configurations, six synchronized 1080p ego-centric views, and rich ground-truth annotations including RGB, camera parameters, depth, semantic and instance segmentation, and static--dynamic masks.
  \item \textbf{A unified benchmark for generative FVV synthesis and efficient delivery.}
  The benchmark supports systematic evaluation of 3DGS and 4DGS novel-view synthesis, 4DGS streaming, and representation compression under a common camera geometry and rendering protocol.
  \item \textbf{Five challenge tracks spanning synthesis, streaming, and compression.}
  We define Track 1 for 3DGS synthesis, Track 2 for 4DGS synthesis, Track 3 for 4DGS streaming, Track 4 for 3DGS compression, and Track 5 for 4DGS compression.
  \item \textbf{A reproducible evaluation protocol.}
  We treat rendering quality, representation rate, computational complexity, and temporal consistency as complementary evaluation dimensions and provide representative baseline results and a feedforward compression reference design for future challenge participants.
\end{itemize}

\section{Related Work}
\subsection{3D Gaussian Splatting}
3D Gaussian Splatting (3DGS)~\cite{kerbl20233d} represents a scene with anisotropic Gaussian primitives parameterized by position, scale, rotation, opacity, and appearance, enabling efficient optimization and real-time rendering. Based on this representation, recent works have explored reconstruction under challenging input conditions, particularly sparse-view settings. CoR-GS~\cite{zhang2024cor}, FSGS~\cite{zhu2024fsgs}, and DropGaussian~\cite{park2025dropgaussian} improve reconstruction quality and efficiency by exploiting geometric priors, adaptive Gaussian optimization, or redundancy reduction. Meanwhile, compact Gaussian representations have been investigated through anchor-based and grid-based structures~\cite{lu2024scaffold,morgenstern2024compact}, aiming to reduce representation size and improve the efficiency of Gaussian organization. These advances have established 3DGS as an efficient foundation for extending Gaussian representations to dynamic scenes.

\subsection{4D Gaussian Splatting}
Building upon 3DGS, 4D Gaussian Splatting extends Gaussian representations to dynamic scenes by modeling the temporal evolution of geometry and appearance. Existing methods primarily introduce deformation fields or time-conditioned transformations to update Gaussian primitives over time~\cite{yang2024deformable,wu20244d}. More recent approaches further explore structured spatio-temporal representations to improve reconstruction efficiency and modeling capacity~\cite{hu20254dgc}. Despite these advances, most existing 4D Gaussian methods rely on offline optimization with access to the complete temporal sequence. This limits their applicability to streaming scenarios, where the scene must be incrementally reconstructed and updated as new frames arrive.
  
\subsection{4D Gaussian Splatting Streaming}
Online reconstruction of dynamic scenes is considerably more challenging than conventional offline training, as the model must continuously update its representation from incoming frames rather than relying on the complete multi-view sequence. Early studies investigated this setting primarily with NeRF-based representations~\cite{mildenhall2021nerf,barron2021mip,barron2022mip,cao2023hexplane,fridovich2023k,muller2022instant}. StreamRF~\cite{li2022streaming} followed an incremental learning strategy by capturing changes on a frame-by-frame basis, whereas ReRF~\cite{wang2023neural} encoded residuals between neighboring frames to enable reconstruction over extended sequences. NeRFPlayer~\cite{song2023nerfplayer}, meanwhile, factorized the 4D spatio-temporal space to achieve a more compact representation and faster reconstruction.
The introduction of 3DGS significantly improved the efficiency of both optimization and rendering, making it particularly suitable for streaming dynamic reconstruction. 3DGStream~\cite{sun20243dgstream} leveraged a Neural Transformation Cache to efficiently update the motion of Gaussian primitives. Subsequent approaches have investigated more sophisticated motion representations, including HiCoM~\cite{gao2024hicom}, which models motion through a hierarchical coherent mechanism, and IGS~\cite{yan2025instant}, which employs an anchor-driven network to estimate motion residuals in a single step, thereby accelerating training. To further balance representation capacity and storage efficiency, QUEEN~\cite{girish2024queen} combines Gaussian residuals with learned quantization and sparsity, enabling a more compact representation for streaming reconstruction.

\subsection{Gaussian Compression}
% Early compression methods reduce the number or precision of Gaussian parameters using pruning, quantization, vector quantization, or compact parameterizations~\cite{fan2024lightgaussian,lee2024compact,niedermayr2024compressed,navaneet2311compact3d,chen20254dgs,balle2018variational,cheng2020learned,minnen2018joint}. More recent methods exploit structure and context: Scaffold-GS~\cite{lu2024scaffold} introduces anchor-based organization, HAC~\cite{chen2024hac} uses hash-grid-assisted context, ContextGS~\cite{wang2024contextgs} models anchor-level context, and mixture-of-prior coding~\cite{liu20253d} improves entropy modeling. Feedforward directions such as FCGS~\cite{chen2025fast} and D-FCGS~\cite{zhang2026d} are particularly relevant to the low-latency objective of this benchmark.
3DGS/4DGS represent scenes with learnable Gaussian primitives, enabling high-quality reconstruction, fast optimization, and real-time rendering. However, the large number of Gaussians and associated attributes incurs considerable storage costs, motivating research on efficient Gaussian compression.

Early approaches mainly reduce parameter redundancy through vector quantization~\cite{fan2023lightgaussian,lee2024compact,navaneet2023compact3d,niedermayr2024compressed} or Gaussian pruning~\cite{fan2023lightgaussian,lee2024compact}. Inspired by the neural image compression~\cite{balle2018variational,minnen2018joint,cheng2020learned,liu2023icmh,liu2025efficient} and the neural point cloud compression~\cite{huang2020octsqueeze,que2021voxelcontext,liu2023icme,liu2024towards}, later studies~\cite{lu2024scaffold,morgenstern2023compact,chen2025hac,wang2024contextgs,chen20254dgs,liu2024hemgs,liu20253d} exploit structural dependencies for more compact representations. For example, Scaffold-GS~\cite{lu2024scaffold} organizes scene features around anchors but does not employ entropy coding. Building upon it, HAC~\cite{chen2025hac} models spatial correlations with a hash grid for entropy coding, while ContextGS~\cite{wang2024contextgs} utilizes anchor-level context as hyperprior information for efficient compression. Further, MoP~\cite{liu20253d} uses a mixture-of-priors approach, compressing the 3DGS in different ways for better compression.

\subsection{Datasets for 3D/4D Gaussian Evaluation}
Static reconstruction commonly uses Mip-NeRF 360~\cite{barron2022mip}, BungeeNeRF~\cite{xiangli2022bungeenerf}, Synthetic-NeRF~\cite{mildenhall2021nerf}, DeepBlending~\cite{hedman2018deep}, Tank\&Temples~\cite{knapitsch2017tanks} and DL3DV-10K~\cite{ling2024dl3dv}. Dynamic multi-view evaluation additionally uses N3DV, Meeting Room, SelfCap, Google Immersive Video, Ego-Exo4D, EPIC-KITCHENS, and HOI4D ~\cite{li2022neural,li2022streaming,xu2024representing,broxton2020immersive,grauman2024ego,damen2020epic,liu2022hoi4d}. \MISR complements these resources by isolating geometry and redundancy factors under synthetic, fully controlled conditions. 

\begin{figure}[t]
  \centering
  \includegraphics[width=\columnwidth]{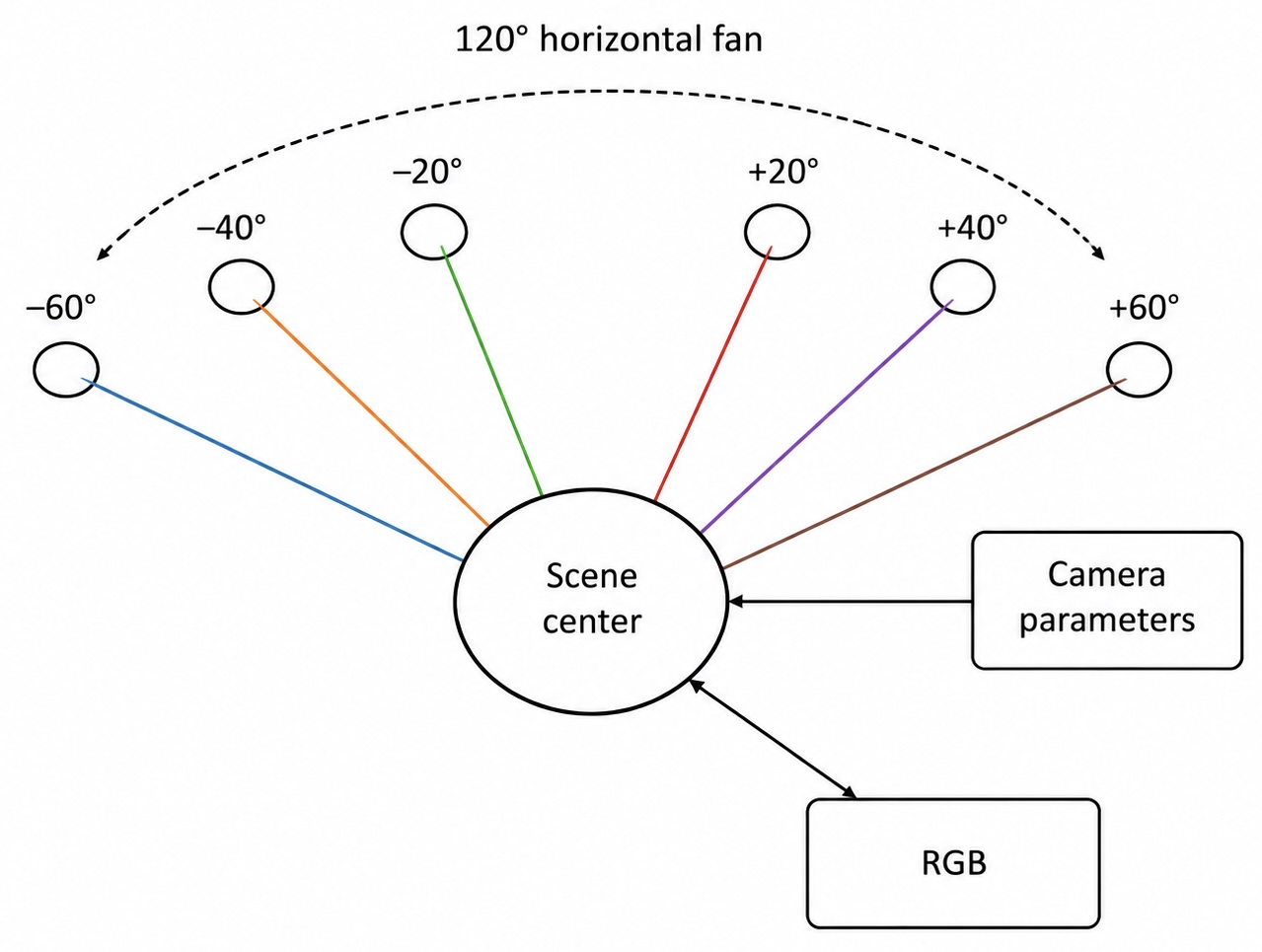}
  \vspace{-2mm}
  \caption{M$^3$ISR camera configuration. Six synchronized views share a common scene center and cover a $120^\circ$ horizontal fan at yaw angles $\{-60^\circ,-40^\circ,-20^\circ,+20^\circ,+40^\circ,+60^\circ\}$. Each frame is paired with RGB and camera parameters.}
  \vspace{-2mm}
  \label{fig:rig}
\end{figure}

\section{M$^3$ISR Dataset}
\subsection{Dataset Overview}
\MISR contains five scene categories: \emph{Bedroom}, \emph{Kitchen}, \emph{LivingRoom}, \emph{Outdoor1}, and \emph{Outdoor2}. Each category contains five scene instances, giving 25 scenes in total. Every scene is rendered in two configurations: \emph{MovingCameraStaticScene}, where the camera moves while the scene content remains static, and \emph{MovingCameraDynamicScene}, where camera motion and object motion occur simultaneously.

Each sequence is rendered at $1920\times1080$ using 1024 samples per pixel. A sequence lasts 2 seconds at 15 FPS, corresponding to 30 timestamps. With 25 scenes, two configurations, and six synchronized views, the dataset contains
\begin{equation}
25\times2\times6\times30 = 9{,}000
\end{equation}
synchronized RGB frames.

In addition to RGB images and calibrated camera parameters, \MISR provides noise-free ground-truth depth, semantic segmentation, instance segmentation, and static--dynamic scene masks. These annotations support not only reconstruction and novel-view synthesis, but also geometry-aware sparsification, content-aware bit allocation, motion-aware representation design, and fine-grained evaluation of spatial and temporal reconstruction quality.

\begin{figure*}[t]
    \centering
    \includegraphics[width=\textwidth]{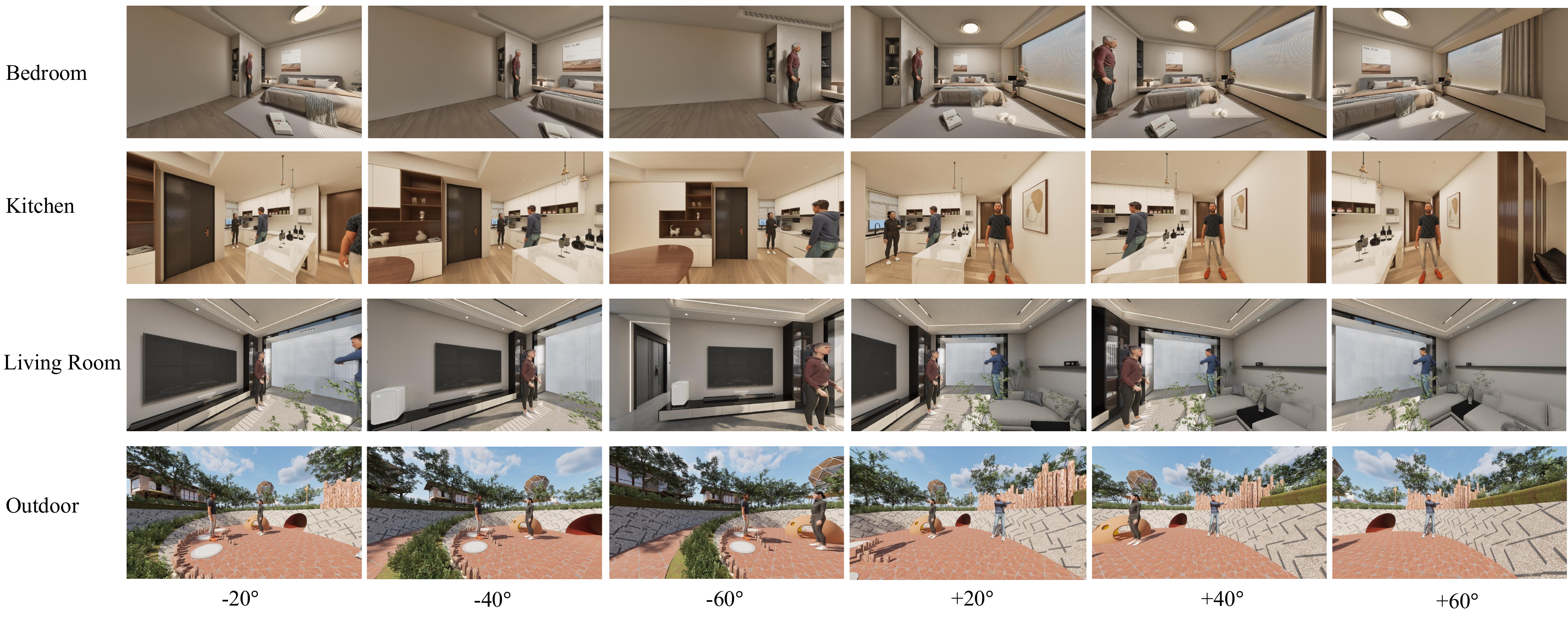}
  \caption{Representative synchronized multi-view samples from \MISR. The displayed figure shows four representative scene groups; the Outdoor row groups the two dataset categories Outdoor1 and Outdoor2 for readability. Each row contains six synchronized camera views.}
    \label{fig:overview}
\end{figure*}

\subsection{Camera Geometry and Evaluation Protocol}
The six synchronized cameras share a common scene center, intentionally removing translational parallax and isolating view-directional variation. Their horizontal yaw angles are
\begin{equation}
\Theta=\{-60^\circ,-40^\circ,-20^\circ,+20^\circ,+40^\circ,+60^\circ\}.
\end{equation}
This yields a $120^\circ$ angular range with $20^\circ$ spacing between adjacent cameras and a $40^\circ$ central gap, providing a controlled setting for angular novel-view synthesis and representation efficiency. The six cameras serve as input views, while additional camera poses are withheld for evaluation. The held-out poses and scene-level split are fixed across methods to ensure reproducibility and prevent test-time leakage. RGB frames are always paired with calibrated camera parameters.

\subsection{Comparison with Representative Dynamic Multi-View Datasets}
Table~\ref{tab:datasets} summarizes representative resources discussed in the supplied benchmark survey. We intentionally omit a definitive frame rate for SelfCap because the provided survey reports inconsistent values (30 FPS in its summary table and 60 FPS in the accompanying description); this should be verified against the official release before camera-ready submission.

\begin{table*}[t]
\centering
\caption{Representative datasets for 3DGS/4DGS streaming evaluation. R denotes real capture and S denotes synthetic. M$^3$ISR provides six synchronized views with a shared-center camera layout.}
\label{tab:datasets}
\resizebox{\textwidth}{!}{%
\begin{tabular}{lcccccl}
\toprule
Dataset & Year/Venue & Type & \#Cams & Resolution & FPS & Scenes / Rig Geometry \\
\midrule
N3DV~\cite{li2022neural} & CVPR 2022 & R & 18--21 & $2028\times2704$ & 30 & 6 scenes; forward-facing planar \\
Meeting Room~\cite{li2022streaming} & NeurIPS 2022 & R & 13 & $1280\times720$ & 30 & 3 scenes; forward-facing planar \\
SelfCap~\cite{xu2024representing} & SIGGRAPH Asia 2024 & R & 22 & $3840\times2160$ & -- & 3 sequences; inward-facing surround \\
Google Immersive Video~\cite{broxton2020immersive} & SIGGRAPH 2020 & R & 46 & $2560\times1920$ & 30 & 15 scenes; hemispherical dome \\
Ego-Exo4D~\cite{grauman2024ego} & CVPR 2024 & R & 1--4 ego + exo & $1440\times1440$ & 30 & large-scale hybrid ego/exo capture \\
EPIC-KITCHENS-100~\cite{damen2020epic} & PAMI 2020 & R & ego (+ exo) & $1920\times1080$ & 60 & egocentric trajectories \\
HOI4D~\cite{liu2022hoi4d} & CVPR 2022 & R & 1 RGB-D & $1280\times720$ & 30 & 2.4M frames; egocentric trajectory \\
\MISR (ours) & 2026 & S & 6 & $1920\times1080$ & 15 & 25 scenes; shared-center $\pm60^\circ$ fan \\
\bottomrule
\end{tabular}}
\end{table*}

\subsection{Why M$^3$ISR for Compression Research?}
\textbf{View-aware bit allocation.} The synchronized camera rig makes it possible to evaluate view-dependent pruning and precision allocation across the six camera directions.
\textbf{Cross-view sparsification.} The fixed camera geometry supports controlled measurements of view-dependent Gaussian pruning and representation compactness.
\textbf{Cross-view redundancy.} The synchronized six-view rig provides a fixed setting for view-dependent importance modeling and bandwidth adaptation.
\textbf{Temporal redundancy.} The paired static/dynamic configurations separate camera-induced view changes from object motion, making temporal reuse easier to study.

\section{Challenge Tracks and Evaluation Protocol}
This Challenge is organized into five complementary tracks covering generative FVV synthesis and efficient FVV delivery. Tracks 1 and 2 focus on novel-view generation, while Tracks 3--5 target efficient representation and delivery under storage, bitrate, and latency constraints.

\subsection{Track 1: 3D Gaussian Splatting}
Track 1 focuses on generative FVV synthesis using static 3D Gaussian Splatting. Given sparse synchronized multi-view observations, participants are expected to reconstruct a compact 3D representation and synthesize novel views with high visual fidelity. The track emphasizes the trade-off between rendering quality and computational efficiency, particularly under the wide-FoV sparse camera geometry provided by \MISR.

We evaluate reconstruction quality using PSNR, SSIM, and LPIPS on held-out views. Representation storage and rendering speed are additionally reported to characterize the practical efficiency of each method.

\subsection{Track 2: 4D Gaussian Splatting}
Track 2 extends generative FVV synthesis to dynamic scenes using 4D Gaussian Splatting and related time-conditioned representations. The objective is to recover high-fidelity novel views while preserving temporal consistency under simultaneous camera and object motion.

In addition to frame-level rendering quality, the track emphasizes the ability of a representation to model spatial and temporal variations jointly. PSNR, SSIM, and LPIPS are evaluated across held-out views and timestamps, while storage and rendering speed are reported to characterize the practical
quality--efficiency trade-off.

\subsection{Track 3: 4D Gaussian Splatting Streaming}
Track 3 targets efficient delivery of 4D Gaussian representations for free-viewpoint video. While 4DGS enables dynamic novel-view synthesis, practical streaming introduces stringent constraints on transmission bandwidth, representation update cost, and end-to-end latency. The objective of this track is therefore to develop streaming strategies that preserve perceptual quality while reducing transmission and reconstruction overhead.

We evaluate visual quality together with representation storage, reported training or reconstruction time, and rendering throughput. We distinguish training time from end-to-end streaming latency: the former measures the computational cost reported by the reference implementation, whereas the latter measures the time from the arrival of the first input frame until the target representation becomes available for rendering. Since end-to-end latency is not available for all current reference baselines, the baseline table reports training or reconstruction time separately and does not treat it as a direct measurement of streaming latency.

\begin{figure*}[htbp]
    \centering
    \includegraphics[width=\textwidth]{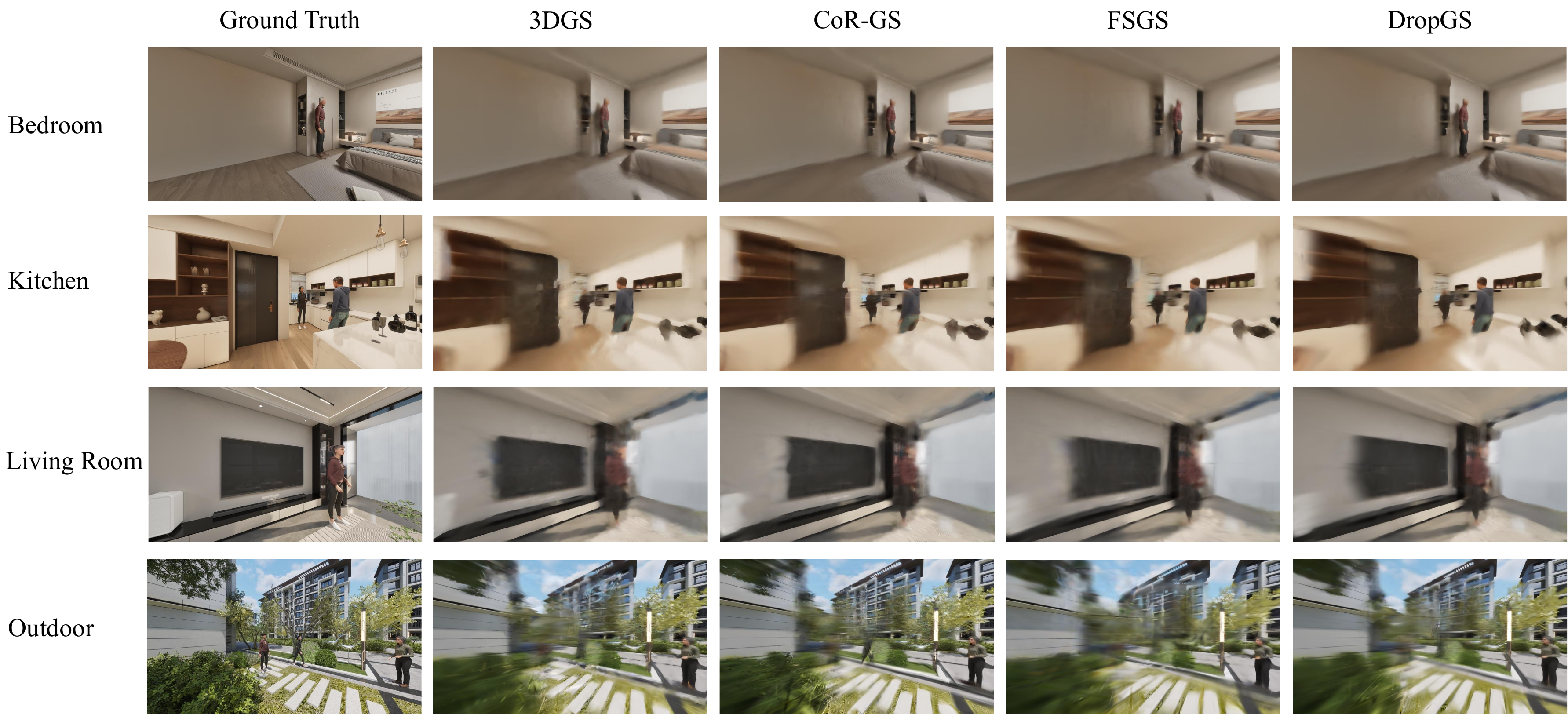}
    \caption{Qualitative comparison of Track 1 sparse-view 3DGS reconstruction on \MISR. Ground truth and results from 3DGS, CoR-GS, FSGS, and DropGS are shown for representative indoor and outdoor scenes.}
\label{fig:sparseview3dgs}
\end{figure*}

\subsection{Track 4: 3D Gaussian Splatting Compression}
Track 4 focuses on reducing the storage and transmission cost of static 3DGS representations while preserving rendering quality. A 3DGS representation
contains a large number of Gaussian primitives and associated attributes, which can introduce substantial redundancy and limit practical deployment.

The goal is to learn compact representations that achieve a favorable trade-off between representation rate, rendering fidelity, and computational
efficiency. When applicable, participants should report multiple operating points to characterize the rate--distortion behavior of their methods.

Given a static Gaussian representation
\begin{equation}
\mathcal{G}=\{\mathbf{x}_i,\mathbf{s}_i,\mathbf{q}_i,\alpha_i,\mathbf{c}_i\}_{i=1}^{N},
\end{equation}
where the variables denote position, scale, rotation, opacity, and appearance, the goal is to produce a compact bitstream from which a decoded representation $\hat{\mathcal{G}}$ yields high-quality rendered views. The defining constraint of Task D is that test-time compression should be a single feedforward pass, with no scene-specific optimization.

\subsection{Task 5: 4D Gaussian Splatting Compression}
Track 5 targets compact compression of dynamic 4D Gaussian representations. Compared with static 3DGS, 4DGS introduces additional redundancy across both space and time, resulting in substantially higher representation complexity. The challenge is to reduce this redundancy while preserving both spatial rendering fidelity and temporal consistency.

The track therefore evaluates the rate--distortion--complexity trade-off of 4DGS compression. In addition to coded representation size and rendering
quality, decoding efficiency and temporal reconstruction quality should be considered when evaluating practical deployment.
\begin{equation}
\mathcal{G}_t = \mathcal{T}(\mathcal{G}_{t-1},\mathbf{m}_t,\mathbf{r}_t),
\end{equation}
where $\mathbf{m}_t$ denotes motion and $\mathbf{r}_t$ denotes rotation and/or residual attributes. Task E evaluates whether a shared feedforward model can compress the reference representation and temporal changes without independently re-optimizing every timestamp.

\section{Recommended Evaluation Metrics}
To provide a consistent evaluation across all five tracks, we consider four complementary dimensions: rendering quality, representation rate, computational complexity, and temporal consistency.
\paragraph{Rendering quality.}
PSNR, SSIM, and LPIPS are computed between rendered results and ground-truth images and averaged over the designated evaluation views and timestamps.
\paragraph{Temporal consistency.}
For dynamic tracks, evaluation should additionally consider temporal stability and consistency across consecutive timestamps, in order to distinguish high per-frame fidelity from temporally coherent FVV synthesis.
\paragraph{Rate and representation cost.}
Reported representation size should include all information required for decoding and rendering, including Gaussian parameters, latent codes, hyperlatents, entropy-model side information, temporal residuals, and required
metadata.
\paragraph{Computational efficiency.}
Encoding time, decoding time, peak memory, and rendering FPS should be reported when applicable. For streaming tracks, end-to-end latency and per-scene encoding or reconstruction time should be measured explicitly.
\paragraph{Rate--distortion behavior.}
For compression and delivery tracks, multiple operating points are encouraged to characterize the trade-off among bitrate or storage, perceptual quality, and computational complexity.

\section{Baseline Evaluation on \MISR}
We provide representative baseline results for Tracks 1--3 to establish a common reference across static 3DGS, 4DGS streaming, and dynamic 4DGS reconstruction. These results are intended to characterize the benchmark difficulty and motivate the five challenge tracks rather than to establish a new state-of-the-art method. Unless otherwise stated, the reported numbers are taken from the supplied benchmark evidence.
We use them to characterize the benchmark, not to claim reimplementation beyond the provided evidence.

\subsection{Static 3DGS Reconstruction}
We evaluate vanilla 3DGS~\cite{kerbl20233d}, CoR-GS~\cite{zhang2024cor}, FSGS~\cite{zhu2024fsgs}, and DropGaussian~\cite{park2025dropgaussian}. Table~\ref{tab:static} reports average results over 25 scenes.

\begin{table}[htbp]
\centering
\caption{Average static 3DGS metrics on \MISR. Storage is total disk usage over 25 scenes.}
\label{tab:static}
\resizebox{\columnwidth}{!}{%
\begin{tabular}{lcccc}
\toprule
Method & PSNR $\uparrow$ & SSIM $\uparrow$ & LPIPS $\downarrow$ & Storage (MB) $\downarrow$ \\
\midrule
3DGS & 18.705 & 0.655 & 0.461 & 1331 \\
CoR-GS & 18.825 & \textbf{0.658} & \textbf{0.445} & 3174 \\
FSGS & 18.726 & 0.654 & 0.475 & \textbf{508} \\
DropGaussian & \textbf{18.864} & 0.657 & 0.466 & 992 \\
\bottomrule
\end{tabular}}
\end{table}

The main observation is not a large gap in rendering quality, but a large gap in storage. PSNR varies by only 0.159 dB between the best and worst entries, while total storage spans 508--3,174 MB. FSGS therefore achieves a compact representation with similar reconstruction quality, whereas CoR-GS spends considerably more storage for a comparatively small PSNR gain over vanilla 3DGS. This behavior indicates substantial representation redundancy under the six-view wide-FoV setting.

\subsection{Streaming Gaussian Reconstruction}
The supplied results for QUEEN~\cite{girish2024queen} and an octree/LoD-structured streaming baseline~\cite{liu2026lod} are summarized in Table~\ref{tab:streaming}.

\begin{table*}[htbp]
\centering
\caption{Average streaming metrics reported on \MISR. Time refers to the training or reconstruction time reported by the reference implementation and should not be interpreted as end-to-end streaming latency.}
\label{tab:streaming}
\begin{tabular}{lcccccc}
\toprule
Method & PSNR $\uparrow$ & SSIM $\uparrow$ & LPIPS $\downarrow$ & Storage (MB) $\downarrow$ & Train/Recon. Time (s) $\downarrow$ & Render FPS $\uparrow$ \\
\midrule
QUEEN~\cite{girish2024queen} & 15.93 & 0.5125 & 0.4786 & 1734 & \textbf{351.48} & 54.64 \\
Octree-/LoD-structured~\cite{liu2026lod} & \textbf{16.32} & \textbf{0.5336} & \textbf{0.4650} & \textbf{1452} & 356.90 & \textbf{60.10} \\
\bottomrule
\end{tabular}
\end{table*}

\begin{figure}[htbp]
    \centering
    \includegraphics[width=0.48\textwidth]{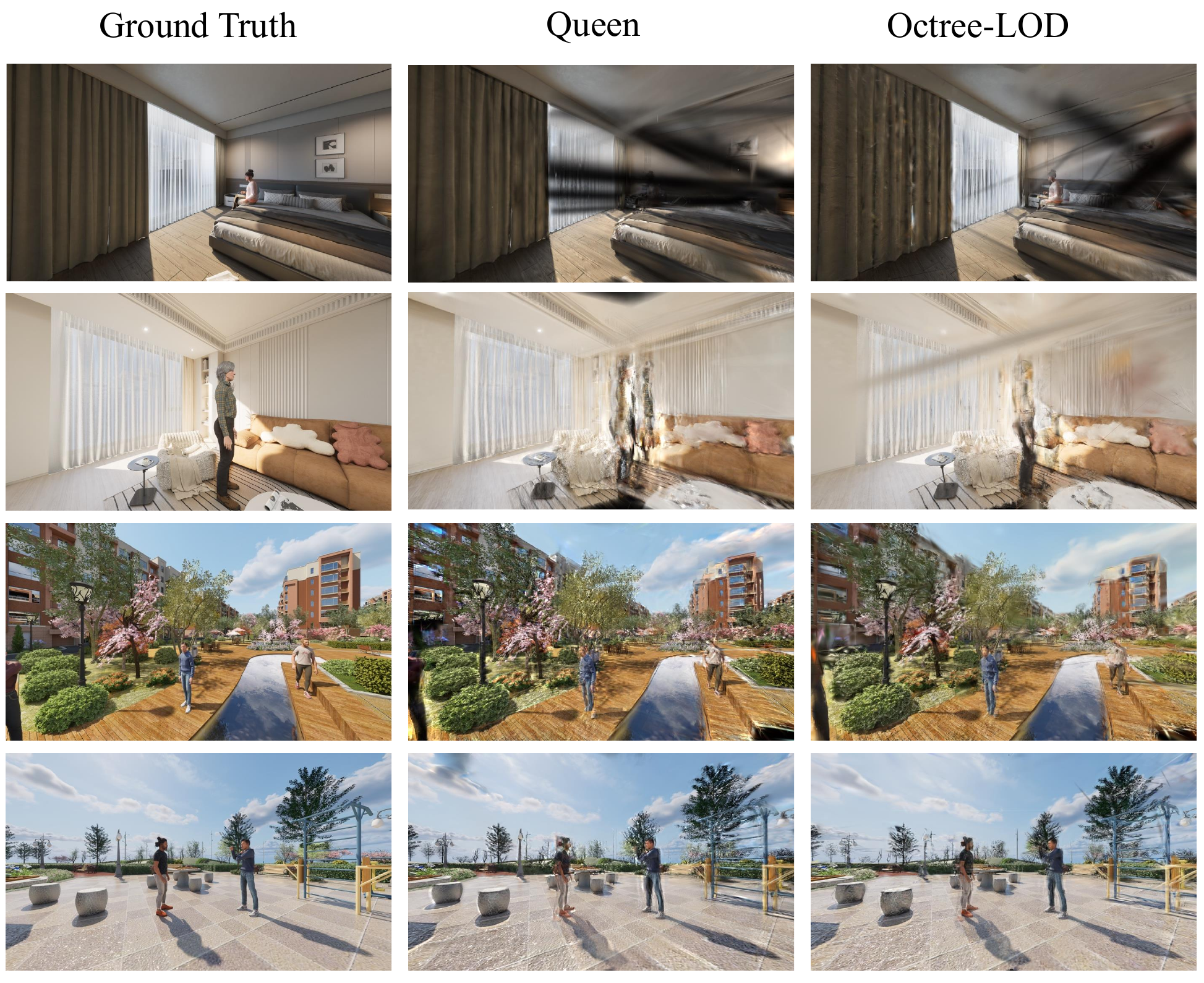}
    \vspace{-2mm}
    \caption{Qualitative comparison of Track 3 4DGS streaming reconstruction on \MISR. Ground truth and results from QUEEN and Octree-LOD are shown for representative dynamic scenes.}
    \vspace{-2mm}
    \label{fig:streaming}
\end{figure}

Both streaming baselines report more than 350 s of training or reconstruction time for a 2 s clip. Their storage also remains high at 1,452--1,734 MB over the 25-scene benchmark. The lower PSNR relative to the static 3DGS baselines suggests that simultaneous camera and object motion remains challenging for the evaluated streaming methods.

\subsection{Dynamic 4DGS Reconstruction}
Table~\ref{tab:4dgs} summarizes the provided results for Deformable 3DGS~\cite{yang2024deformable} and 4DGS~\cite{wu20244d}.

\begin{table}[htbp]
\centering
\caption{Average dynamic Gaussian reconstruction results on \MISR.}
\vspace{-2mm}
\label{tab:4dgs}
\resizebox{\columnwidth}{!}{%
\begin{tabular}{lccccc}
\toprule
Method & PSNR $\uparrow$ & SSIM $\uparrow$ & LPIPS $\downarrow$ & Storage (MB) $\downarrow$ & FPS $\uparrow$ \\
\midrule
Deformable 3DGS~\cite{yang2024deformable} & \textbf{22.79} & 0.7258 & 0.4024 & 257 & 12.84 \\
4DGS~\cite{wu20244d} & 22.57 & \textbf{0.7323} & \textbf{0.3598} & \textbf{200} & \textbf{52.84} \\
\bottomrule
\end{tabular}}
\vspace{-2mm}
\end{table}

\begin{figure}[htbp]
    \centering
    \includegraphics[width=0.48\textwidth]{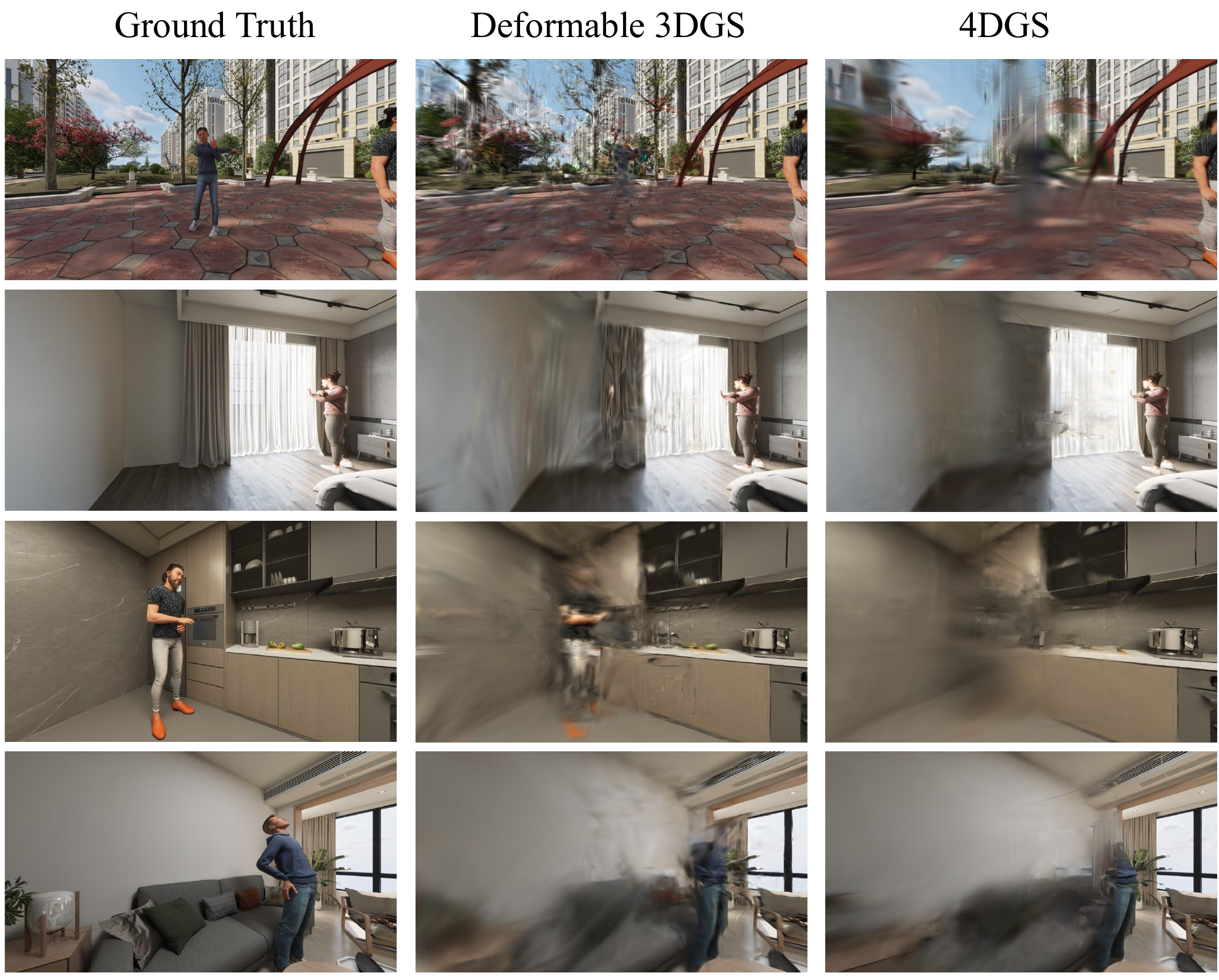}
    \vspace{-2mm}
    \caption{Qualitative comparison of Track 2 dynamic 4DGS reconstruction on \MISR. Ground truth and results from Deformable 3DGS and 4DGS are shown for representative dynamic scenes.}
    \label{fig:4DGS}
    \vspace{-2mm}
\end{figure}

Compared with the streaming baselines, the two time-conditioned methods substantially improve both fidelity and compactness. 4DGS achieves 22.57 dB PSNR at 200 MB while rendering at 52.84 FPS; Deformable 3DGS reaches slightly higher PSNR at the cost of both storage and rendering speed. The key benchmark-level observation is that explicit spatio-temporal modeling is more effective than treating each frame as an independent reconstruction. Nevertheless, 200--257 MB over 25 scenes still corresponds to approximately 8--10 MB per 2 s scene sequence, leaving clear room for compression.

\subsection{Feedforward 3D Gaussian Splatting Compression}

3D Gaussian Splatting provides an explicit representation for high-quality novel-view synthesis. However, the large number of Gaussian primitives and their associated attributes introduce substantial storage and transmission costs, limiting practical deployment in immersive applications. Existing compression methods have explored pruning, quantization, vector quantization, and structural parameterization to reduce redundancy. Nevertheless, many existing approaches still rely on scene-specific optimization, which increases encoding latency and limits scalability.

The objective of Track 4 is to develop efficient compression methods for static 3DGS representations. Given a Gaussian representation
\begin{equation}
\mathcal{G}=\{x_i,s_i,q_i,\alpha_i,c_i\}_{i=1}^{N},
\end{equation}
where $x_i$, $s_i$, $q_i$, $\alpha_i$, and $c_i$ denote Gaussian position, scale, rotation, opacity, and appearance attributes, respectively. The goal for 3DGS compression is to generate a compact coded representation, from which a decoded Gaussian model
$\hat{\mathcal{G}}$ can reconstruct high-quality novel views.

% The key constraint of this track is that compression should be performed using a feedforward inference process without per-scene optimization. 
To optimize the compression framework, methods should learn transferable priors across scenes and achieve a favorable trade-off among representation rate, rendering fidelity, and computational complexity.
A generic rate--distortion formulation can be written as
\begin{equation}
\mathcal{L}_{3D}=R(\hat{\mathbf{y}})+R(\hat{\mathbf{z}})+\lambda D(\mathcal{G},\hat{\mathcal{G}}),
\end{equation}
where $\mathbf{y}$ and $\mathbf{z}$ denote latent representations and hyperprior variables, respectively. A compression framework consists of an analysis transform, hyperprior modeling, quantization, entropy coding, and a synthesis transform:
\begin{equation}
\mathbf{y}=g_a(\mathcal{G}),\qquad
\mathbf{z}=h_a(\mathbf{y}),
\end{equation}
\begin{equation}
\hat{\mathbf{y}}=Q(\mathbf{y}),\qquad
\hat{\mathbf{z}}=Q(\mathbf{z}),
\end{equation}
followed by reconstruction
\begin{equation}
\hat{\mathcal{G}}=g_s(\hat{\mathbf{y}}).
\end{equation}
Participants are encouraged to explore compact Gaussian parameterization, learned entropy models, structural redundancy reduction, and view-aware allocation strategies.

% Evaluation should consider both compression efficiency and rendering quality. The reported metrics should include coded representation size, rendering quality (PSNR, SSIM, LPIPS), decoding complexity, and rendering throughput. Multiple operating points are recommended to characterize the rate--distortion--complexity trade-off.

\begin{figure}[t!]
    \centering
    \includegraphics[width=\linewidth]{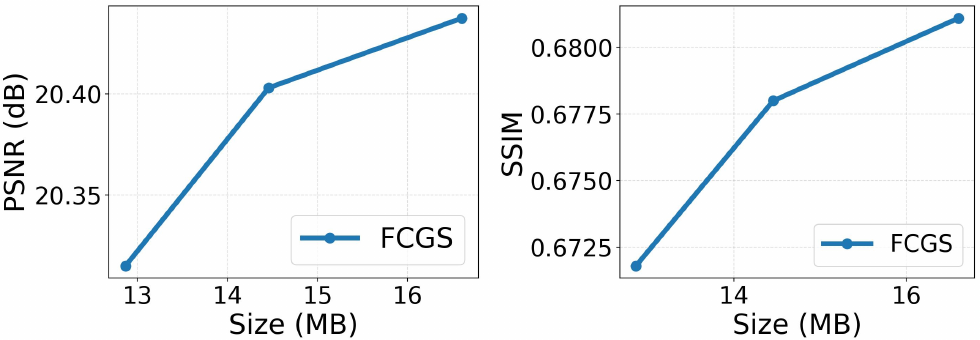}
    \vspace{-2mm}
    \caption{The 3DGS compression performance of the FCGS method on our \MISR benchmark.}
    \vspace{-2mm}
    \label{fig:FCGS}
\end{figure}

\begin{figure}[t!]
    \centering
    \includegraphics[width=\linewidth]{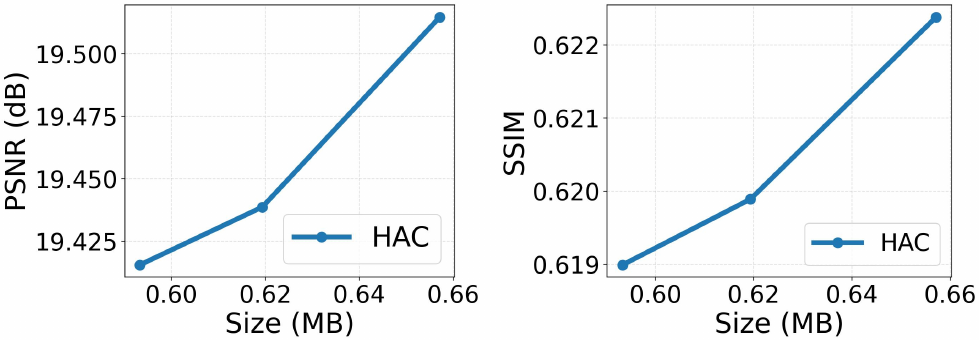}
    \vspace{-2mm}
    \caption{The 3DGS compression performance of the HAC method on our \MISR benchmark.}\vspace{-2mm}
    \label{fig:HAC}
\end{figure}

Here, we evaluate our benchmark \MISR using a feed-forward compression method (\textit{i.e.}, FCGS~\cite{fcgs2025}). We report both PSNR and SSIM to provide a comprehensive evaluation of compression performance. The results in Figure~\ref{fig:FCGS} demonstrate that our benchmark effectively reflects the rate-distortion (R-D) trade-off of feed-forward compression methods. Additionally, optimization-based compression methods also play an important role in 3DGS compression. To further demonstrate the generalizability of our benchmark to optimization-based methods, we evaluate their compression performance on our \MISR, as shown in Figure~\ref{fig:HAC}. The resulting R-D curves further demonstrate that \MISR can effectively evaluate the compression performance of optimization-based methods, highlighting its applicability across different compression paradigms.

\subsection{Feedforward 4D Gaussian Splatting Compression}
4D Gaussian Splatting extends static Gaussian representations into dynamic scenes by modeling spatial and temporal variations jointly. While this enables high-quality free-viewpoint video synthesis, it also introduces substantial redundancy across both spatial primitives and temporal updates. Efficiently compressing these dynamic representations while preserving temporal coherence remains an open challenge.

The objective of Track 5 is to develop feedforward compression approaches for 4DGS representations that reduce storage and transmission cost while maintaining spatial fidelity, temporal consistency, and efficient decoding.

A dynamic Gaussian representation can be modeled as
\begin{equation}
\mathcal{G}_{t} = T(\mathcal{G}_{t-1},\mathbf{m}_{t},\mathbf{r}_{t}),
\end{equation}
where $\mathbf{m}_{t}$ represents motion-related information and $\mathbf{r}_{t}$ represents rotation or additional residual attributes. Instead of independently encoding every timestamp, efficient approaches should exploit temporal redundancy by separating static reference information from dynamic residual components.

For example, motion and residual attributes can be transformed into latent representations:
\begin{equation}
\mathbf{y}^{m}_{t}=g^{m}_{a}(\mathbf{m}_{t}), \mathbf{y}^{r}_{t}= g^{r}_{a}(\mathbf{r}_{t}).
\end{equation}
The temporal rate can then be formulated as
\begin{equation}
R_t = R(\hat{\mathbf{y}}^{m}_{t}) + R(\hat{\mathbf{y}}^{r}_{t}) + R(\hat{\mathbf{z}}_{t}),
\end{equation}
with the overall optimization objective
\begin{equation}
\mathcal{L}_{4D} = \sum_t (R_t+\lambda D_t).
\end{equation}

\begin{figure}[t!]
    \centering
    \includegraphics[width=\linewidth]{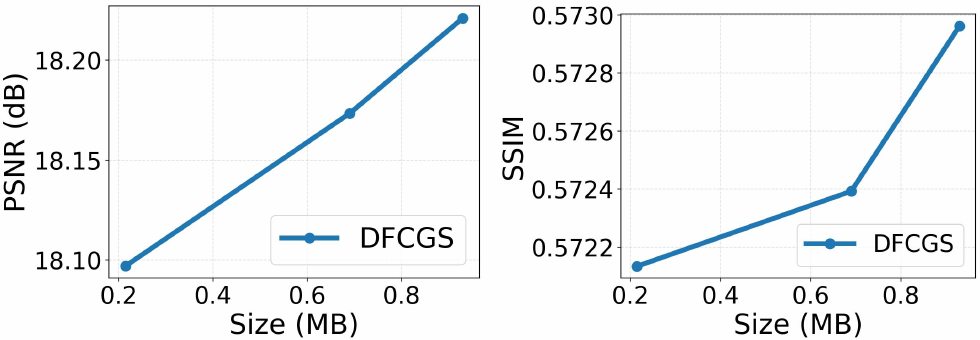}
    \caption{The 4DGS compression performance of the D-FCGS method on our \MISR benchmark.}
    \label{fig:DFCGS}
\end{figure}

We evaluate our \MISR using the feedforward compression method D-FCGS~\cite{zhang2026d}, as shown in Figure~\ref{fig:DFCGS}. Although D-FCGS is trained exclusively on real-world datasets, it still achieves favorable compression performance when directly evaluated on our synthetically generated \MISR without additional adaptation. This result suggests that \MISR exhibits performance characteristics consistent with those of real-world datasets, despite being fully synthetic. Such consistency further demonstrates effectiveness and applicability of \MISR for evaluating 4DGS compression methods.

\subsection{Compression Baselines and Evaluation Considerations}
The benchmark survey includes reference rate--quality comparisons for existing Gaussian compression approaches such as HAC, FCGS, and D-FCGS. However, the available descriptions do not provide a complete machine-readable specification
of all operating points, coded-size definitions, scene splits, rendering protocols, and encoding-time measurements. Therefore, these curves are treated as motivation for Tracks 4 and 5 rather than directly comparable quantitative evidence for a new M3ISR compression method.
A rigorous leaderboard should require all participants to report:
\begin{itemize}
    \item exact coded representation size, including latent codes, entropy-model side information, metadata, and temporal residuals when applicable;
    \item rendering protocol and evaluation views;
    \item training and encoding time measurement;
    \item decoding complexity and memory consumption;
    \item rate--distortion curves across multiple operating points.
\end{itemize}
Such a protocol enables fair comparison between feedforward compression models and optimization-based approaches, and provides a reproducible evaluation framework for future Gaussian representation compression research.

\section{Discussion and Limitations}
\textbf{Controlled benchmark versus realism.} \MISR is intentionally synthetic. Its fixed camera rig and rendering pipeline remove several failure modes present in real captures, including calibration error, motion blur, exposure variation, and sensor noise. The benchmark should therefore complement rather than replace real-world validation.
\textbf{Sparse-view geometry.} The six-camera $120^\circ$ fan is useful for stress-testing wide-view extrapolation, but it is not a substitute for full $360^\circ$ immersive rigs such as Google Immersive Video. Future benchmark extensions should vary the number of views and camera baseline while keeping scene identity fixed.
\textbf{Sequence length.} Each sequence lasts only 2 seconds. This is sufficient for short-horizon temporal redundancy analysis but cannot characterize long-term drift, memory growth, or scene changes over minutes. The long-horizon behavior should therefore be assessed on resources such as SelfCap in addition to \MISR.
\textbf{Benchmark scope.}
The current baseline evidence reveals substantial differences in rendering quality, representation storage, and computational cost, but it does not by itself establish a complete end-to-end streaming latency profile or the performance of a new feedforward codec. The reference compression formulations should therefore be interpreted as challenge specifications and reproducibility-oriented starting points. A completed benchmark evaluation should additionally report held-out scenes, multiple rate points, end-to-end latency, and cross-dataset generalization.

\section{Conclusion}
We presented \MISR, a controlled multi-view benchmark for evaluating generative FVV synthesis and efficient neural delivery with 3D and 4D Gaussian representations. The benchmark provides 25 synthetic indoor and
outdoor scenes, two static/dynamic configurations, six synchronized ego-centric 1080p views, calibrated camera parameters, and dense ground-truth annotations.
Our benchmark is organized into five complementary challenge tracks covering 3DGS synthesis, 4DGS synthesis, 4DGS streaming, 3DGS compression, and 4DGS compression. Representative baseline results reveal substantial differences in rendering fidelity, storage efficiency, temporal reconstruction, and computational cost, motivating research that jointly considers quality, representation rate, and system efficiency.
More broadly, \MISR is designed as a reproducible testbed for the full FVV pipeline, from novel-view generation to compact representation and low-latency delivery. We hope the benchmark and challenge protocol will facilitate
systematic comparison and encourage new methods for view-aware, temporal, and rate-efficient Gaussian representations.

\bibliographystyle{ACM-Reference-Format}
\bibliography{sample-base}

@String{Computing = "Computing" }

@String{Computer = "{IEEE} Computer" }

@String{Springer = "Springer-Verlag" }

@String(CVPR= {IEEE Conf. Comput. Vis. Pattern Recog.})

@String(ECCV= {Eur. Conf. Comput. Vis.})

@String(TOG= {ACM Trans. Graph.})

@String(ICME = {Int. Conf. Multimedia and Expo})

@String(AAAI = {AAAI})

@String(CVPR  = {CVPR})

@String(ECCV  = {ECCV})

@String(TOG   = {ACM TOG})

@String(ICME  =	{ICME})

@ArtifactSoftware{R,
    title = {R: A Language and Environment for Statistical Computing},
    author = {{R Core Team}},
    organization = {R Foundation for Statistical Computing},
    address = {Vienna, Austria},
    year = {2019},
    url = {https://www.R-project.org/},
}

@article{mildenhall2021nerf,
  title={Nerf: Representing scenes as neural radiance fields for view synthesis},
  author={Mildenhall, Ben and Srinivasan, Pratul P and Tancik, Matthew and Barron, Jonathan T and Ramamoorthi, Ravi and Ng, Ren},
  journal={Communications of the ACM},
  volume={65},
  number={1},
  pages={99--106},
  year={2021},
  publisher={ACM New York, NY, USA}
}

@article{kerbl20233d,
  title={3d gaussian splatting for real-time radiance field rendering.},
  author={Kerbl, Bernhard and Kopanas, Georgios and Leimk{\"u}hler, Thomas and Drettakis, George and others},
  journal={ACM Trans. Graph.},
  volume={42},
  number={4},
  pages={139--1},
  year={2023}
}

@inproceedings{yang2024deformable,
  title={Deformable 3d gaussians for high-fidelity monocular dynamic scene reconstruction},
  author={Yang, Ziyi and Gao, Xinyu and Zhou, Wen and Jiao, Shaohui and Zhang, Yuqing and Jin, Xiaogang},
  booktitle={2024 IEEE/CVF Conference on Computer Vision and Pattern Recognition (CVPR)},
  pages={20331--20341},
  year={2024},
  organization={IEEE}
}

@inproceedings{wu20244d,
  title={4d gaussian splatting for real-time dynamic scene rendering},
  author={Wu, Guanjun and Yi, Taoran and Fang, Jiemin and Xie, Lingxi and Zhang, Xiaopeng and Wei, Wei and Liu, Wenyu and Tian, Qi and Wang, Xinggang},
  booktitle={2024 IEEE/CVF Conference on Computer Vision and Pattern Recognition (CVPR)},
  pages={20310--20320},
  year={2024},
  organization={IEEE}
}

@article{girish2024queen,
  title={Queen: Quantized efficient encoding of dynamic gaussians for streaming free-viewpoint videos},
  author={Girish, Sharath and Li, Tianye and Mazumdar, Amrita and Shrivastava, Abhinav and Luebke, David and De Mello, Shalini},
  journal={Advances in Neural Information Processing Systems},
  volume={37},
  pages={43435--43467},
  year={2024}
}

@article{liu2026lod,
  title={LoD-Structured 3D Gaussian Splatting for Streaming Video Reconstruction},
  author={Liu, Xinhui and Wang, Can and Liu, Lei and Chen, Zhenghao and Jiang, Wei and Wang, Wei and Xu, Dong},
  journal={arXiv preprint arXiv:2601.18475},
  year={2026}
}

@inproceedings{sun20243dgstream,
  title={3dgstream: On-the-fly training of 3d gaussians for efficient streaming of photo-realistic free-viewpoint videos},
  author={Sun, Jiakai and Jiao, Han and Li, Guangyuan and Zhang, Zhanjie and Zhao, Lei and Xing, Wei},
  booktitle={2024 IEEE/CVF Conference on Computer Vision and Pattern Recognition (CVPR)},
  pages={20675--20685},
  year={2024},
  organization={IEEE}
}

@inproceedings{lee2024compact,
  title={Compact 3d gaussian representation for radiance field},
  author={Lee, Joo Chan and Rho, Daniel and Sun, Xiangyu and Ko, Jong Hwan and Park, Eunbyung},
  booktitle={2024 IEEE/CVF Conference on Computer Vision and Pattern Recognition (CVPR)},
  pages={21719--21728},
  year={2024},
  organization={IEEE}
}

@inproceedings{niedermayr2024compressed,
  title={Compressed 3d gaussian splatting for accelerated novel view synthesis},
  author={Niedermayr, Simon and Stumpfegger, Josef and Westermann, R{\"u}diger},
  booktitle={2024 IEEE/CVF Conference on Computer Vision and Pattern Recognition (CVPR)},
  pages={10349--10358},
  year={2024},
  organization={IEEE}
}

@inproceedings{lu2024scaffold,
  title={Scaffold-gs: Structured 3d gaussians for view-adaptive rendering},
  author={Lu, Tao and Yu, Mulin and Xu, Linning and Xiangli, Yuanbo and Wang, Limin and Lin, Dahua and Dai, Bo},
  booktitle={2024 IEEE/CVF Conference on Computer Vision and Pattern Recognition (CVPR)},
  pages={20654--20664},
  year={2024},
  organization={IEEE}
}

@article{wang2024contextgs,
  title={Contextgs: Compact 3d gaussian splatting with anchor level context model},
  author={Wang, Yufei and Li, Zhihao and Guo, Lanqing and Yang, Wenhan and Kot, Alex C and Wen, Bihan},
  journal={Advances in neural information processing systems},
  volume={37},
  pages={51532--51551},
  year={2024}
}

@inproceedings{liu20253d,
  title={3d gaussian splatting data compression with mixture of priors},
  author={Liu, Lei and Chen, Zhenghao and Xu, Dong},
  booktitle={Proceedings of the 33rd ACM International Conference on Multimedia},
  pages={8341--8350},
  year={2025}
}

@inproceedings{hu20254dgc,
  title={4dgc: Rate-aware 4d gaussian compression for efficient streamable free-viewpoint video},
  author={Hu, Qiang and Zheng, Zihan and Zhong, Houqiang and Fu, Sihua and Song, Li and Zhang, Xiaoyun and Zhai, Guangtao and Wang, Yanfeng},
  booktitle={2025 IEEE/CVF Conference on Computer Vision and Pattern Recognition (CVPR)},
  pages={875--885},
  year={2025},
  organization={IEEE}
}

@article{chen20254dgs,
  title={4dgs-cc: A contextual coding framework for 4d gaussian splatting data compression},
  author={Chen, Zicong and Chen, Zhenghao and Jiang, Wei and Wang, Wei and Liu, Lei and Xu, Dong},
  journal={arXiv preprint arXiv:2504.18925},
  year={2025}
}

@inproceedings{zhang2026d,
  title={D-FCGS: Feedforward compression of dynamic Gaussian splatting for free-viewpoint videos},
  author={Zhang, Wenkang and Zhao, Yan and Wang, Qiang and Xu, Zhixin and Song, Li and Cheng, Zhengxue},
  booktitle={Proceedings of the AAAI Conference on Artificial Intelligence},
  volume={40},
  number={19},
  pages={16361--16369},
  year={2026}
}

@article{balle2018variational,
  title={Variational image compression with a scale hyperprior},
  author={Ball{\'e}, Johannes and Minnen, David and Singh, Saurabh and Hwang, Sung Jin and Johnston, Nick},
  journal={arXiv preprint arXiv:1802.01436},
  year={2018}
}

@inproceedings{cheng2020learned,
  title={Learned image compression with discretized gaussian mixture likelihoods and attention modules},
  author={Cheng, Zhengxue and Sun, Heming and Takeuchi, Masaru and Katto, Jiro},
  booktitle={2020 IEEE/CVF Conference on Computer Vision and Pattern Recognition (CVPR)},
  pages={7936--7945},
  year={2020},
  organization={IEEE}
}

@article{minnen2018joint,
  title={Joint autoregressive and hierarchical priors for learned image compression},
  author={Minnen, David and Ball{\'e}, Johannes and Toderici, George D},
  journal={Advances in neural information processing systems},
  volume={31},
  year={2018}
}

@inproceedings{zhang2024cor,
  title={Cor-gs: sparse-view 3d gaussian splatting via co-regularization},
  author={Zhang, Jiawei and Li, Jiahe and Yu, Xiaohan and Huang, Lei and Gu, Lin and Zheng, Jin and Bai, Xiao},
  booktitle={European conference on computer vision},
  pages={335--352},
  year={2024},
  organization={Springer}
}

@inproceedings{zhu2024fsgs,
  title={Fsgs: Real-time few-shot view synthesis using gaussian splatting},
  author={Zhu, Zehao and Fan, Zhiwen and Jiang, Yifan and Wang, Zhangyang},
  booktitle={European conference on computer vision},
  pages={145--163},
  year={2024},
  organization={Springer}
}

@inproceedings{park2025dropgaussian,
  title={Dropgaussian: Structural regularization for sparse-view gaussian splatting},
  author={Park, Hyunwoo and Ryu, Gun and Kim, Wonjun},
  booktitle={2025 IEEE/CVF Conference on Computer Vision and Pattern Recognition (CVPR)},
  pages={21600--21609},
  year={2025},
  organization={IEEE}
}

@inproceedings{morgenstern2024compact,
  title={Compact 3d scene representation via self-organizing gaussian grids},
  author={Morgenstern, Wieland and Barthel, Florian and Hilsmann, Anna and Eisert, Peter},
  booktitle={European Conference on Computer Vision},
  pages={18--34},
  year={2024},
  organization={Springer}
}

@inproceedings{barron2022mip,
  title={Mip-nerf 360: Unbounded anti-aliased neural radiance fields},
  author={Barron, Jonathan T and Mildenhall, Ben and Verbin, Dor and Srinivasan, Pratul P and Hedman, Peter},
  booktitle={2022 IEEE/CVF Conference on Computer Vision and Pattern Recognition (CVPR)},
  pages={5460--5469},
  year={2022},
  organization={IEEE}
}

@inproceedings{xiangli2022bungeenerf,
  title={Bungeenerf: Progressive neural radiance field for extreme multi-scale scene rendering},
  author={Xiangli, Yuanbo and Xu, Linning and Pan, Xingang and Zhao, Nanxuan and Rao, Anyi and Theobalt, Christian and Dai, Bo and Lin, Dahua},
  booktitle={European conference on computer vision},
  pages={106--122},
  year={2022},
  organization={Springer}
}

@inproceedings{ling2024dl3dv,
  title={Dl3dv-10k: A large-scale scene dataset for deep learning-based 3d vision},
  author={Ling, Lu and Sheng, Yichen and Tu, Zhi and Zhao, Wentian and Xin, Cheng and Wan, Kun and Yu, Lantao and Guo, Qianyu and Yu, Zixun and Lu, Yawen and others},
  booktitle={2024 IEEE/CVF Conference on Computer Vision and Pattern Recognition (CVPR)},
  pages={22160--22169},
  year={2024},
  organization={IEEE}
}

@inproceedings{li2022neural,
  title={Neural 3d video synthesis from multi-view video},
  author={Li, Tianye and Slavcheva, Mira and Zollhoefer, Michael and Green, Simon and Lassner, Christoph and Kim, Changil and Schmidt, Tanner and Lovegrove, Steven and Goesele, Michael and Newcombe, Richard and others},
  booktitle={2022 IEEE/CVF Conference on Computer Vision and Pattern Recognition (CVPR)},
  pages={5511--5521},
  year={2022},
  organization={IEEE}
}

@article{li2022streaming,
  title={Streaming radiance fields for 3d video synthesis},
  author={Li, Lingzhi and Shen, Zhen and Wang, Zhongshu and Shen, Li and Tan, Ping},
  journal={Advances in Neural Information Processing Systems},
  volume={35},
  pages={13485--13498},
  year={2022}
}

@article{xu2024representing,
  title={Representing long volumetric video with temporal gaussian hierarchy},
  author={Xu, Zhen and Xu, Yinghao and Yu, Zhiyuan and Peng, Sida and Sun, Jiaming and Bao, Hujun and Zhou, Xiaowei},
  journal={ACM Transactions on Graphics (TOG)},
  volume={43},
  number={6},
  pages={1--18},
  year={2024},
  publisher={ACM New York, NY, USA}
}

@article{broxton2020immersive,
  title={Immersive light field video with a layered mesh representation},
  author={Broxton, Michael and Flynn, John and Overbeck, Ryan and Erickson, Daniel and Hedman, Peter and Duvall, Matthew and Dourgarian, Jason and Busch, Jay and Whalen, Matt and Debevec, Paul},
  journal={ACM Transactions on Graphics (TOG)},
  volume={39},
  number={4},
  pages={86--1},
  year={2020},
  publisher={ACM New York, NY, USA}
}

@inproceedings{grauman2024ego,
  title={Ego-exo4d: Understanding skilled human activity from first-and third-person perspectives},
  author={Grauman, Kristen and Westbury, Andrew and Torresani, Lorenzo and Kitani, Kris and Malik, Jitendra and Afouras, Triantafyllos and Ashutosh, Kumar and Baiyya, Vijay and Bansal, Siddhant and Boote, Bikram and others},
  booktitle={Proceedings of the IEEE/CVF conference on computer vision and pattern recognition},
  pages={19383--19400},
  year={2024}
}

@article{damen2020epic,
  title={The epic-kitchens dataset: Collection, challenges and baselines},
  author={Damen, Dima and Doughty, Hazel and Farinella, Giovanni Maria and Fidler, Sanja and Furnari, Antonino and Kazakos, Evangelos and Moltisanti, Davide and Munro, Jonathan and Perrett, Toby and Price, Will and others},
  journal={IEEE Transactions on Pattern Analysis and Machine Intelligence},
  volume={43},
  number={11},
  pages={4125--4141},
  year={2020},
  publisher={IEEE}
}

@inproceedings{liu2022hoi4d,
  title={Hoi4d: A 4d egocentric dataset for category-level human-object interaction},
  author={Liu, Yunze and Liu, Yun and Jiang, Che and Lyu, Kangbo and Wan, Weikang and Shen, Hao and Liang, Boqiang and Fu, Zhoujie and Wang, He and Yi, Li},
  booktitle={2022 IEEE/CVF Conference on Computer Vision and Pattern Recognition (CVPR)},
  pages={20981--20990},
  year={2022},
  organization={IEEE}
}

@inproceedings{barron2021mip,
  title={Mip-nerf: A multiscale representation for anti-aliasing neural radiance fields},
  author={Barron, Jonathan T and Mildenhall, Ben and Tancik, Matthew and Hedman, Peter and Martin-Brualla, Ricardo and Srinivasan, Pratul P},
  booktitle={Proceedings of the IEEE/CVF international conference on computer vision},
  pages={5855--5864},
  year={2021}
}

@inproceedings{cao2023hexplane,
  title={Hexplane: A fast representation for dynamic scenes},
  author={Cao, Ang and Johnson, Justin},
  booktitle={Proceedings of the IEEE/CVF Conference on Computer Vision and Pattern Recognition},
  pages={130--141},
  year={2023}
}

@inproceedings{fridovich2023k,
  title={K-planes: Explicit radiance fields in space, time, and appearance},
  author={Fridovich-Keil, Sara and Meanti, Giacomo and Warburg, Frederik Rahb{\ae}k and Recht, Benjamin and Kanazawa, Angjoo},
  booktitle={Proceedings of the IEEE/CVF Conference on Computer Vision and Pattern Recognition},
  pages={12479--12488},
  year={2023}
}

@article{muller2022instant,
  title={Instant neural graphics primitives with a multiresolution hash encoding},
  author={M{\"u}ller, Thomas and Evans, Alex and Schied, Christoph and Keller, Alexander},
  journal={ACM transactions on graphics (TOG)},
  volume={41},
  number={4},
  pages={1--15},
  year={2022},
  publisher={ACM New York, NY, USA}
}

@inproceedings{wang2023neural,
  title={Neural residual radiance fields for streamably free-viewpoint videos},
  author={Wang, Liao and Hu, Qiang and He, Qihan and Wang, Ziyu and Yu, Jingyi and Tuytelaars, Tinne and Xu, Lan and Wu, Minye},
  booktitle={Proceedings of the IEEE/CVF Conference on Computer Vision and Pattern Recognition},
  pages={76--87},
  year={2023}
}

@article{song2023nerfplayer,
  title={Nerfplayer: A streamable dynamic scene representation with decomposed neural radiance fields},
  author={Song, Liangchen and Chen, Anpei and Li, Zhong and Chen, Zhang and Chen, Lele and Yuan, Junsong and Xu, Yi and Geiger, Andreas},
  journal={IEEE Transactions on Visualization and Computer Graphics},
  volume={29},
  number={5},
  pages={2732--2742},
  year={2023},
  publisher={IEEE}
}

@article{gao2024hicom,
  title={Hicom: Hierarchical coherent motion for dynamic streamable scenes with 3d gaussian splatting},
  author={Gao, Qiankun and Meng, Jiarui and Wen, Chengxiang and Chen, Jie and Zhang, Jian},
  journal={Advances in Neural Information Processing Systems},
  volume={37},
  pages={80609--80633},
  year={2024}
}

@inproceedings{yan2025instant,
  title={Instant gaussian stream: Fast and generalizable streaming of dynamic scene reconstruction via gaussian splatting},
  author={Yan, Jinbo and Peng, Rui and Wang, Zhiyan and Tang, Luyang and Yang, Jiayu and Liang, Jie and Wu, Jiahao and Wang, Ronggang},
  booktitle={Proceedings of the Computer Vision and Pattern Recognition Conference},
  pages={16520--16531},
  year={2025}
}

@article{navaneet2023compact3d,
  title={Compact3d: Compressing gaussian splat radiance field models with vector quantization},
  author={Navaneet, KL and Meibodi, Kossar Pourahmadi and Koohpayegani, Soroush Abbasi and Pirsiavash, Hamed},
  journal={arXiv preprint arXiv:2311.18159},
  year={2023}
}

@article{fan2023lightgaussian,
  title={Lightgaussian: Unbounded 3d gaussian compression with 15x reduction and 200+ fps},
  author={Fan, Zhiwen and Wang, Kevin and Wen, Kairun and Zhu, Zehao and Xu, Dejia and Wang, Zhangyang},
  journal={arXiv preprint arXiv:2311.17245},
  year={2023}
}

@INPROCEEDINGS{liu2023icme,
  author={Liu, Lei and Hu, Zhihao and Zhang, Jing},
  booktitle={2023 IEEE International Conference on Multimedia and Expo (ICME)}, 
  title={PCHM-Net: A New Point Cloud Compression Framework for Both Human Vision and Machine Vision}, 
  year={2023},
  volume={},
  number={},
  pages={1997-2002}
}

@inproceedings{liu2024towards,
  title={Towards point cloud compression for machine perception: A simple and strong baseline by learning the octree depth level predictor},
  author={Liu, Lei and Hu, Zhihao and Chen, Zhenghao},
  booktitle={International Joint Conference on Artificial Intelligence WorkShop},
  pages={3--17},
  year={2024},
  organization={Springer}
}

@article{liu2025efficient,
  title={An Efficient Adaptive Compression Method for Human Perception and Machine Vision Tasks},
  author={Liu, Lei and Chen, Zhenghao and Hu, Zhihao and Xu, Dong},
  journal={Pattern Recognition},
  year={2026}
}

@inproceedings{liu2023icmh,
  title={Icmh-net: Neural image compression towards both machine vision and human vision},
  author={Liu, Lei and Hu, Zhihao and Chen, Zhenghao and Xu, Dong},
  booktitle={Proceedings of the 31st ACM International Conference on Multimedia},
  pages={8047--8056},
  year={2023}
}

@inproceedings{que2021voxelcontext,
  title={Voxelcontext-net: An octree based framework for point cloud compression},
  author={Que, Zizheng and Lu, Guo and Xu, Dong},
  booktitle={2021 IEEE/CVF Conference on Computer Vision and Pattern Recognition (CVPR)},
  pages={6038--6047},
  year={2021},
  organization={IEEE}
}

@inproceedings{huang2020octsqueeze,
  title={Octsqueeze: Octree-structured entropy model for lidar compression},
  author={Huang, Lila and Wang, Shenlong and Wong, Kelvin and Liu, Jerry and Urtasun, Raquel},
  booktitle={2020 IEEE/CVF Conference on Computer Vision and Pattern Recognition (CVPR)},
  pages={1310--1320},
  year={2020},
  organization={IEEE}
}

@article{morgenstern2023compact,
  title={Compact 3d scene representation via self-organizing gaussian grids},
  author={Morgenstern, Wieland and Barthel, Florian and Hilsmann, Anna and Eisert, Peter},
  journal=ECCV,
  year={2024}
}

@inproceedings{chen2025hac,
  title={Hac: Hash-grid assisted context for 3d gaussian splatting compression},
  author={Chen, Yihang and Wu, Qianyi and Lin, Weiyao and Harandi, Mehrtash and Cai, Jianfei},
  booktitle=ECCV,
  pages={422--438},
  year={2024},
  organization={Springer}
}

@article{liu2024hemgs,
  title={Hemgs: A hybrid entropy model for 3d gaussian splatting data compression},
  author={Liu, Lei and Chen, Zhenghao and Jiang, Wei and Wang, Wei and Xu, Dong},
  journal={arXiv preprint arXiv:2411.18473},
  year={2024}
}

@article{hedman2018deep,
  title={Deep blending for free-viewpoint image-based rendering},
  author={Hedman, Peter and Philip, Julien and Price, True and Frahm, Jan-Michael and Drettakis, George and Brostow, Gabriel},
  journal={ACM Trans. Graph.},
  volume={37},
  number={6},
  pages={1--15},
  year={2018},
  publisher={ACM New York, NY, USA}
}

@article{knapitsch2017tanks,
  title={Tanks and temples: Benchmarking large-scale scene reconstruction},
  author={Knapitsch, Arno and Park, Jaesik and Zhou, Qian-Yi and Koltun, Vladlen},
  journal=TOG,
  volume={36},
  number={4},
  pages={1--13},
  year={2017},
  publisher={ACM New York, NY, USA}
}

@inproceedings{fcgs2025,
  title={Fast Feedforward 3D Gaussian Splatting Compression},
  author={Chen, Yihang and Wu, Qianyi and Li, Mengyao and Lin, Weiyao and Harandi, Mehrtash and Cai, Jianfei},
  booktitle={The Thirteenth International Conference on Learning Representations},
  year={2025}
}


\end{document}